\documentclass[runningheads]{llncs}
\usepackage[T1]{fontenc}
\usepackage{graphicx}
\usepackage{booktabs}
\usepackage{hyperref}
\usepackage{util}
\usepackage{multicol}
\usepackage{multirow}
\usepackage{subfig} 
\usepackage{siunitx} 
\usepackage[misc]{ifsym}
\usepackage{amsmath}   
\usepackage{amsfonts}

\usepackage{mwe}
\usepackage{caption}

\begin{document}

\title{Performance is not All You Need: Sustainability Considerations for Algorithms}

\titlerunning{Sustainability Considerations for Algorithms}

\author{Xiang Li\inst{2}$^{,*}$ \and
Chong Zhang\inst{1}$^{,*}$ \and
Hongpeng Wang\inst{4} \and
Shreyank Narayana Gowda\inst{3} \and
Yushi Li\inst{1} \and
Xiaobo Jin\inst{1}$^{,\dagger}$
\institute{Xi'an Jiaotong-Liverpool University \and The Chinese University of Hong Kong \and University of Nottingham \and University of Sydney 
}}

\authorrunning{X. Li, C. Zhang et al.}

\maketitle 

\begin{abstract}
This work focuses on the high carbon emissions generated by deep learning model training, specifically addressing the core challenge of balancing algorithm performance and energy consumption. It proposes an innovative two-dimensional sustainability evaluation system. Different from the traditional single performance-oriented evaluation paradigm, this study pioneered two quantitative indicators that integrate energy efficiency ratio and accuracy: the sustainable harmonic mean (FMS) integrates accumulated energy consumption and performance parameters through the harmonic mean to reveal the algorithm performance under unit energy consumption; the area under the sustainability curve (ASC) constructs a performance-power consumption curve to characterize the energy efficiency characteristics of the algorithm throughout the cycle. To verify the universality of the indicator system, the study constructed benchmarks in various multimodal tasks, including image classification, segmentation, pose estimation, and batch and online learning. Experiments demonstrate that the system can provide a quantitative basis for evaluating cross-task algorithms and promote the transition of green AI research from theory to practice. Our sustainability evaluation framework code can be found here\footnote{$^{*}$ Equal contributions. \\$^{\dagger}$ Corresponding author: Xiaobo Jin \email{\{xiaobo.jin@xjtlu.edu.cn\}}  \\ Code available: \hyperlink{https://github.com/lxgem/NotOnlyPerformence-main/tree/main}{https://github.com/lxgem/NotOnlyPerformence-main/tree/main}}, providing methodological support for the industry to establish algorithm energy efficiency standards.
\end{abstract}


\section{Introduction}
\label{sec:intro}

Recent advances in deep learning have significantly elevated the capabilities of object detection, image classification, and natural language processing, with widespread applications in industrial automation, medical diagnosis, and smart city development~\cite{faiz2023llmcarbon,pouyanfar2018survey}. However, these technological breakthroughs are accompanied by a surge in computational demands, leading to substantial increases in energy consumption. In particular, mainstream deep learning models such as Transformer~\cite{transformer}, ViT~\cite{ViT}, BiLSTM~\cite{BiLSTM}, attention mechanism models~\cite{vaswani2017attention}, and the classic ResNet (Residual Network)~\cite{ResNet} deliver outstanding performance across various tasks but incur high computational complexity and energy costs, escalating hardware requirements and exacerbating environmental impacts.

Current research primarily focuses on optimizing algorithm performance, proposing diverse evaluation metrics to compare algorithms from multiple perspectives~\cite{rainio2024evaluation}. For instance, in image classification, binary tasks employ metrics such as accuracy, recall, and F1-score to address class imbalance, whereas multi-class tasks rely primarily on accuracy. In image segmentation and detection, researchers often employ the Rand index, F1-score, Intersection-over-Union (IoU), Dice coefficient, Average Precision (AP), and Mean Average Precision (mAP). However, these metrics overlook algorithm energy consumption, a critical factor as AI increasingly penetrates edge computing and the demand for green computing grows. The limitations of existing frameworks lie in the dimensional disparity between performance and energy metrics, hindering direct comparisons, and in traditional separate evaluations that may obscure algorithmic advantages within specific energy efficiency ranges. This raises a pivotal scientific question: how can we establish a unified quantitative framework to accurately characterize algorithm energy efficiency while ensuring performance, thereby guiding the development of green AI?

To address this challenge, we propose two novel evaluation metrics to assess the sustainability of algorithms systematically. First, an exponential transformation function maps accumulated energy consumption to a [0,1] interval, aligning it dimensionally with traditional performance metrics (e.g., accuracy, F1-score), enabling the construction of the Sustainable F-measure (FMS) via the harmonic mean of energy and performance metrics. Second, we introduce the sustainability curve framework, which constructs a performance-energy consumption curve through discrete sampling points and computes the Area under the Sustainability Curve (ASC) using numerical integration to form a multidimensional metric. Experiments demonstrate that FMS and ASC effectively balance performance and energy efficiency across tasks like image classification, instance segmentation, pose estimation, and online/batch learning scenarios, providing developers with a quantitative basis for achieving Pareto optimality in performance-energy trade-offs.

\noindent Our main contributions include:
\begin{itemize}
\item We propose a harmonic mean of energy consumption and performance measures to evaluate the sustainability of computing, referred to as the \textbf{F-Measure on Sustainability (FMS)}.
\item With the performance versus power consumption curve, we propose a new sustainability metric based on the area under the curve for sustainability assessment, called \textbf{Area under Sustainability Curve (ASC)}.
\item We verify the rationality of two metrics in multiple tasks, such as image classification, image segmentation, posture recognition, online and batch, and show that they can effectively evaluate the trade-off between energy efficiency and performance of different tasks and models, which have broad applicability.
\end{itemize}

\section{Related work}

\subsection{Measurement on Energy Consumption}

In the field of deep learning, although research on model energy consumption is still relatively limited, some measurement methods and tools have begun to focus on the quantitative evaluation of power consumption. Existing energy evaluation tools, such as Energy-Estimation Toolkits and PowerAPI \cite{leclerc2015powerapi}, can effectively measure the energy performance of deep learning models on different hardware platforms. CodeCarbon \cite{henderson2021codecarbon} streamlines this process by automatically monitoring carbon emissions in machine learning experiments, enabling researchers to gain a deeper understanding of the environmental impact of their models. Similarly, CarbonTracker \cite{anthony2021carbontracker} provides real-time carbon emissions tracking based on computing hardware and regional power grid carbon intensity. GreenAI \cite{schwartz2020green} integrates energy consumption metrics into model development to encourage sustainable practices in deep learning. In addition, MLCO2 \cite{MLCO2} provides a lightweight solution to estimate CO$_2$ emissions using runtime and energy data. By leveraging these tools, researchers can optimize their models to effectively balance performance and energy efficiency.

\ssn{Evaluation Criteria on Algorithms}
There are mainly classification, regression, and ranking tasks in machine learning \cite{bishop2006pattern}. For classification tasks, researchers mainly use accuracy to measure the performance of multi-classification tasks. For two-class tasks, Precision, Recall, F1 metrics, ROC and AUC \footnote{https://en.wikipedia.org/wiki/Evaluation\_of\_binary\_classifiers} are used, where precision represents the proportion of the positive class in all samples predicted to be positive, recall represents the proportion of the positive class in all samples that were originally positive, and F1 is the harmonic mean of the two. In addition, the ROC curve is a graph of the true positive rate (TPR) and the false positive rate (FPR) at each threshold setting, and the AUC is obtained by calculating the area under the ROC curve. For regression tasks, there are Mean Absolute Error (MAE), Mean Squared Error (MSE), Root Mean Squared Error (RMSE), R Squared (R${^2}$), etc. Finally, the researchers use Mean average precision (MAP), DCG and NDCG \cite{jin2008approximately}; Mean reciprocal rank, Precision@n, NDCG@n \footnote{https://en.wikipedia.org/wiki/Learning\_to\_rank}, where $n$ denotes that the metrics are evaluated only on top n documents, etc. to measure the ranking quality of ranking learning. None of the above methods considers the energy consumed by the algorithm. Following the F1 and AUC criteria, we propose two new evaluation criteria that comprehensively consider the performance and energy consumption of the algorithm, which we refer to as sustainability metrics.

\section{Proposed metrics}

To address critical gaps in the evaluation of sustainability-aware algorithms, we propose two novel metrics that systematically combine computational performance with energy efficiency. This dual-metric framework is based on three fundamental requirements: \textbf{1)} the need for dimensionless comparisons across heterogeneous systems, \textbf{2)} mathematical compatibility with traditional performance metrics, and \textbf{3)} practical interpretability for real-world deployment decisions.

\ssn{F-Measure on Sustainability (FMS)}
We propose a new F-measure on sustainability (FMS), which is the harmonic mean of the two metrics of performance on the test set and the power of the training process. FMS achieves a compromise between the two goals of high performance and low power consumption, making our evaluation of machine learning algorithms more objective and reasonable.

First, we use the exponential function to convert the power value to the same scale range as the performance metric, that is, between $(0,1)$. We know that the exponential function has been widely used to model the decay of material life cycles over time. In FMS, the exponential function represents that the energy consumption metric $E(w)$ decreases as the energy consumption $w$ increases. The energy consumption metric  $E(w)$ is defined as follows.
\begin{equation}{\label{eqn:energy}}
    E(w) = e^{-\alpha w},
\end{equation}
where $w$ represents the amount of electricity (with kilowatt per hour, kWh) consumed by the machine learning algorithm, and $\alpha$ is used to control how quickly the energy consumption metric $E(w)$ decays as electricity consumption increases. The power consumption metric has the following obvious properties
\im{
\item The metric is a monotonically decreasing function of the power value, so the trend of the metric value is opposite to the trend of the power value.
\item When the power value $w$ tends to $+\infty$, the function $E(w)$ tends to the minimum value 0; and when the power value $w$ tends to 0, the function $E(w)$ tends to the maximum value 1. No machine learning algorithm will meet these two conditions in the real world.
}

Now we define the $F_{\beta}$-measure on sustainability as the harmonic mean of the energy consumption metric $E$ and the performance metric $P$ ($\beta$ is usually set to $1$)
\begin{equation}\label{eqn:fms}
\tm{FMS} = (1 + \beta^2) \cdot \frac{P \cdot E}{\beta^2 P + E}.
\end{equation}
The harmonic mean intentionally penalizes extreme imbalances: a system achieving $P = 0.9$ with $E = 0.1$ yields FMS = 0.18, whereas balanced $P = E = 0.5$ gives FMS = 0.5.  This enforces sustainable development where energy efficiency and accuracy must co-evolve.

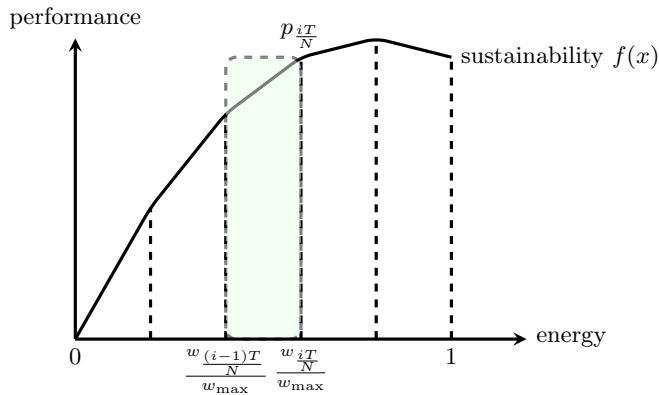
\begin{figure}[!htp]
\vspace{-10pt}
\centering
\begin{tikzpicture}[scale=1.0]
\draw[dashed] (0,0) -- (0,0);
\draw[dashed] (1,0) -- (1,1.75);
\draw[dashed] (2,0) -- (2,3);
\draw[dashed] (3,0) -- (3,3.75);
\draw[dashed] (4,0) -- (4,4);
\draw[dashed] (5,0) -- (5,3.75);
\draw (0,0) -- (1,1.75) -- (2,3) -- (3,3.75) -- (4,4) -- (5,3.75);
\draw (0,0) node[below] {$0$};
\draw (2,0) node[below] {$\frac{w_{\frac{(i - 1)T}{N}}}{w_{\max}}$};
\draw (3,0) node[below] {$\frac{w_{\frac{iT}{N}}}{w_{\max}}$};
\draw (5,0) node[below] {$1$};
\draw[dashed,fill=green!10,opacity=0.5] (2,0) rectangle (3,3.75);
\draw[right] (5,3.75) node {sustainability $f(x)$};
\draw[above] (3,3.75) node {$p_{\frac{iT}{N}}$};
\draw[->] (0,0) -- (0,4) node[above] {performance};
\draw[->] (0,0) -- (6,0) node[right] {energy};
\end{tikzpicture}
\caption{Area under sustainability curve: Energy consumption is normalized to the interval [0,1], and the area under the curve can be calculated by sampling $T$ pairs of samples}\label{fig:ASC}
\vspace{-15pt}
\end{figure}

\ssn{Area under Sustainability Curve (ASC)}
Recognizing that optimal operating points vary across different deployment scenarios, we propose ASC as a comprehensive analyzer of energy-performance tradeoffs.  Deep learning algorithms usually use stochastic gradient optimization algorithms to optimize model parameters. During the iteration process, if we record the energy consumption of training process and performance values on the testset when the algorithm iterates to the $k$-th time, we will get a series of energy consumption and performance value pairs $\{(w_i,p_i)\}_{i = 1}^T$. Then, we can draw a curve (called a sustainability curve) similar to ROC (Receiver Operating Characteristic), which shows the performance value changing with the energy consumption value, as shown in Fig. \ref{fig:ASC}.

Next, we use the rectangular method to calculate the area under the sustainability curve. Assuming the algorithm iterates T times, and the total energy consumption value is $w_{\max}$, we can divide the energy consumption interval and the performance interval into $N$ parts, where each interval contains the same number of iterations. We will get $N$ energy consumption intervals $[w_{(i - 1)T/N}, w_{iT/N}]$ and $N$ performance intervals $[p_{(i - 1)T/N}, p_{iT/N}]$, where $i = 1,2,\cdots,N$. Then, the area under the sustainability curve (ASC) can be approximately calculated as:
\eqn{}{
\tm{ASC} = \int_{0}^1 f(w) \td{w} \approx \sum_{i = 1}^N \frac{w_{iT/N} - w_{(i - 1)T/N}}{w_{\max}}p_{iT/N},
}
where $f$ represents the performance value as a function of the energy consumption value, called the sustainability function, and the energy consumption values at the $t$th iteration are normalized to between $(0,1)$.

Using adaptive Simpson integration for continuous approximation, ASC quantifies an algorithm's ability to maintain performance across energy regimes. Higher ASC values indicate robust systems that preserve accuracy under strict energy constraints—a critical requirement for edge deployment.

\subsection{Scale Invariance of Metrics on Electricity}

Next, we discuss an important property of our measure: scale invariance to power values, which means that both metrics can produce reasonable results for algorithm comparisons on small or large-scale data or models.

For the FMS method, assuming that we have determined the parameter $\alpha_1$ when the algorithm's energy consumption is 1 kWh, then if the large model consumes 1000 kWh on large data, then according to Eqn. (\ref{eqn:energy}), We take $\alpha_2 = \alpha_1/1000$ to maintain identical energy consumption metrics. Therefore, FMS can make energy consumption with different levels scale invariant.

It is worth noting that ASC needs to normalize the power value when calculating the area under the sustainability curve. Assuming that for algorithm A and algorithm B, $w_{\max} = 1$kWh and $w_{\max} = 1000$kWh, then the point $w_i = 0.8$ kWh on algorithm A will correspond to the point $w_j = 800$ kWh on algorithm B, because these two points have the same weight.

In summary, based on the scale invariance of the metric, we can establish a mapping relationship between power levels at different scales. It is worth noting that the performance value here can not only utilize performance metrics, such as accuracy, but also any other performance value in deep learning or machine learning.

\section{Experimental Analysis}
All experiments are conducted using the NVIDIA GeForce RTX 4090 to ensure sufficient computing resources and the repeatability of the results. For the parameter $\alpha$ in the FMS algorithm, it is set to 100 times the energy consumption (kWh) value at the 100th batch in the image classification and online/batch classification applications. It is set to the energy consumed at the 1000th iteration for the other two applications. For the sake of fairness, we consider the power consumed by algorithm training when the performance on the test set is optimal and use this value to calculate the FMS according to Eqn. (\ref{eqn:fms}). Regarding the ASC measure, when the power consumption reaches 1 kWh, we immediately exit the algorithm training process. We evaluate our method in comparison with the following three approaches
\im{
 \item Energy Efficiency Score \cite{score}:
\begin{equation}
\tm{Score} = \frac{P}{E},
\label{eq:score}
\end{equation}
\item Sustainability Index (SI) \cite{fan2023deep}:
\eqn{}{
\tm{SI} = P^{\alpha}  \left( \frac{1}{E} \right)^{\beta},
}
where $\alpha + \beta = 1$ and $\alpha = \beta = 0.5$ generally.
\item Sustainable AI Metric (SAM) \cite{gowda2023watt}:
\begin{equation}
\tm{SAM} = \beta \times \frac{P^{\alpha}}{\log_{10}(E)},
\label{eq:sam}
\end{equation}
where $\alpha$ and $\beta$ are hyperparameters in the criterion, and $\alpha = \beta = 5$.
}
\subsection{Image Classification Task}

\noindent \textbf{Experimental Settings} In the classification task, we experimented with 10\% of the sampled ImageNet \cite{deng2009imagenet} classification dataset to further test the model's ability under controlled sample size, with a batch size of 32. We split the data into 80\% for the training set and 20\% for the test set. We used 3 models of different levels and 2 datasets of different sizes to study the rationality of our evaluation criteria to evaluate the sustainability of different models and different data, including the results of EfficientNet, SwinTransformer\cite{liu2021swin}, and GcViT on ImageNet and CIFAR \cite{cifar}(Tab. \ref{tab:classification_results}). To ensure fairness, all models used are set to the base-level configuration. The lightweight EfficientNet-B0 \cite{tan2019efficientnet} optimizes the network structure to reduce computational complexity and is designed explicitly for resource-constrained environments. In contrast, the GcViT ~\cite{wilkey2020gcvit} model prioritizes high complexity and precision, focusing on accuracy in scenarios that require complex performance.

\begin{table*}[!htp]
\centering
\small
\renewcommand{\arraystretch}{1.1} 
\setlength{\tabcolsep}{4pt} 
\captionsetup{font=footnotesize} 
\vspace{-15pt}
\caption{Comparison of three algorithms on ImageNet and CIFAR in terms of evaluation metrics, parameters (M), and energy consumption.}
\label{tab:classification_results}
\resizebox{\textwidth}{!}{ 
\begin{tabular}{llp{1.5cm}p{1.5cm}p{1.2cm}p{1.2cm}p{1.2cm}p{1.2cm}p{1.2cm}p{1.2cm}}
\toprule
\textbf{Dataset} & \textbf{Model} & \textbf{Params (M)} & \textbf{TE (kWh)} & \textbf{Acc (\%)} & \textbf{Score} & \textbf{SI} & \textbf{SAM} & \textbf{FMS (\%)} & \textbf{ASC (\%)} \\
\midrule
\multirow{3}{*}{\textbf{ImageNet}} 
    & EfficientNet    & \textbf{5.29}  & \textbf{0.73}  & 70.28  & \textbf{0.96} & \textbf{0.98}  & -674.52  & 36.82  & 65.02  \\
    & SwinTransformer & 87.77  & 1.13  & 84.30  & 0.75  & 0.86  &  11.41   & \textbf{39.20}  & \textbf{77.72} \\
    & GcViT           & 90.32  & 1.43  & \textbf{90.40}  & 0.63  & 0.80  & \textbf{13.32}  & 26.58  & 77.02  \\
\midrule
\multirow{3}{*}{\textbf{CIFAR}}    
    & EfficientNet    & \textbf{5.29}  & \textbf{0.99}  & 71.77  & 0.72  & \textbf{0.85}  & -239.81  & 69.45  & 60.11  \\
    & SwinTransformer & 87.77  & 1.42  & \textbf{89.03}  & \textbf{0.79}  & 0.79  & \textbf{21.16}  & \textbf{70.01}  & \textbf{81.24} \\
    & GcViT           & 90.32  & 1.50  & 84.01  & 0.56  & 0.75  & 15.74   & 60.47  & 66.34  \\
\bottomrule
\end{tabular}
}
\vspace{-10pt}
\end{table*}

\begin{figure}[ht]
    \centering
    \includegraphics[width=0.65\textwidth]{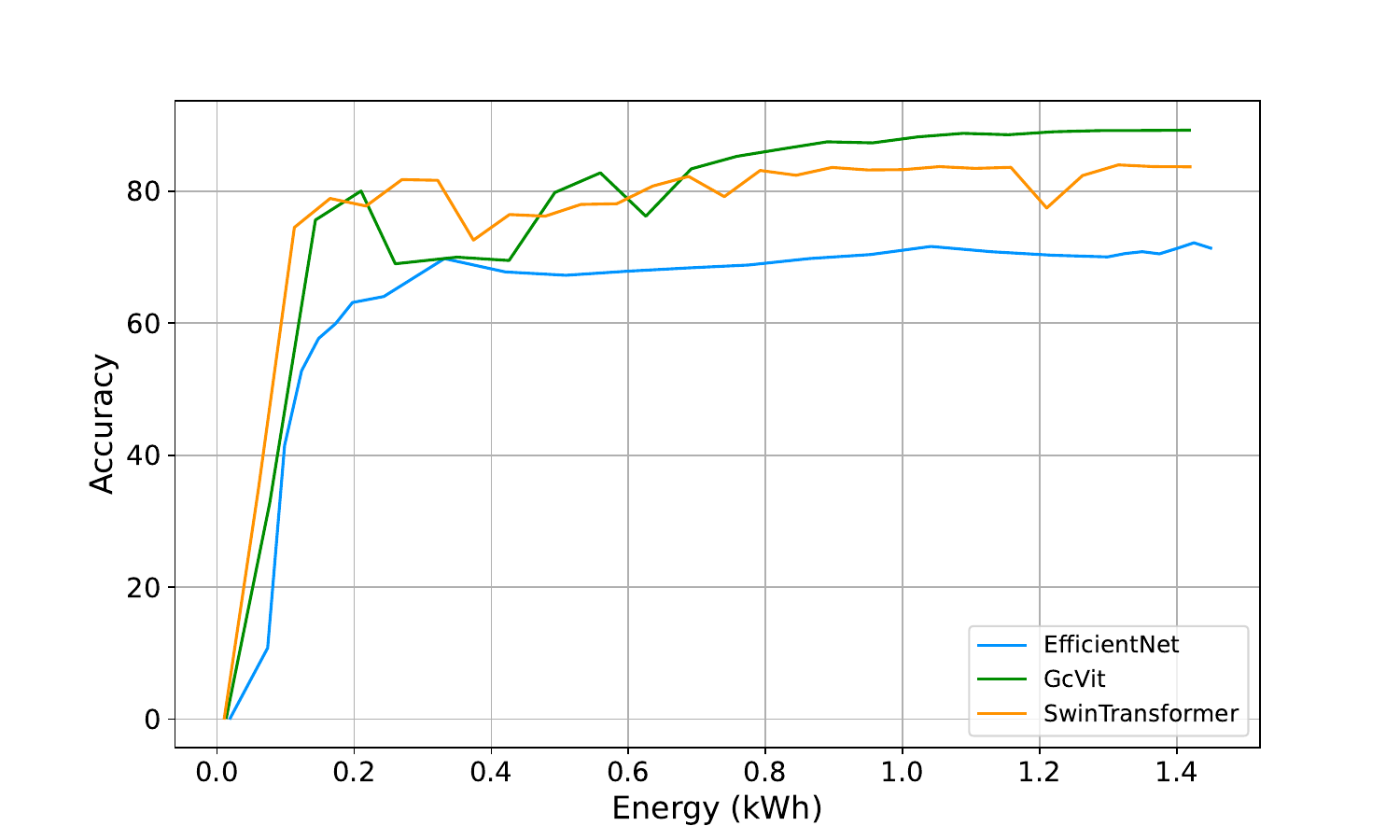}
    \caption{Accuracy vs Energy on ImageNet for EfficientNet, SwinTransformer, and GcViT.}
    \label{fig:accuracy_energy}
    \vspace{-10pt}
\end{figure}

\begin{figure}[ht]
    \centering
    \includegraphics[width=0.65\textwidth]{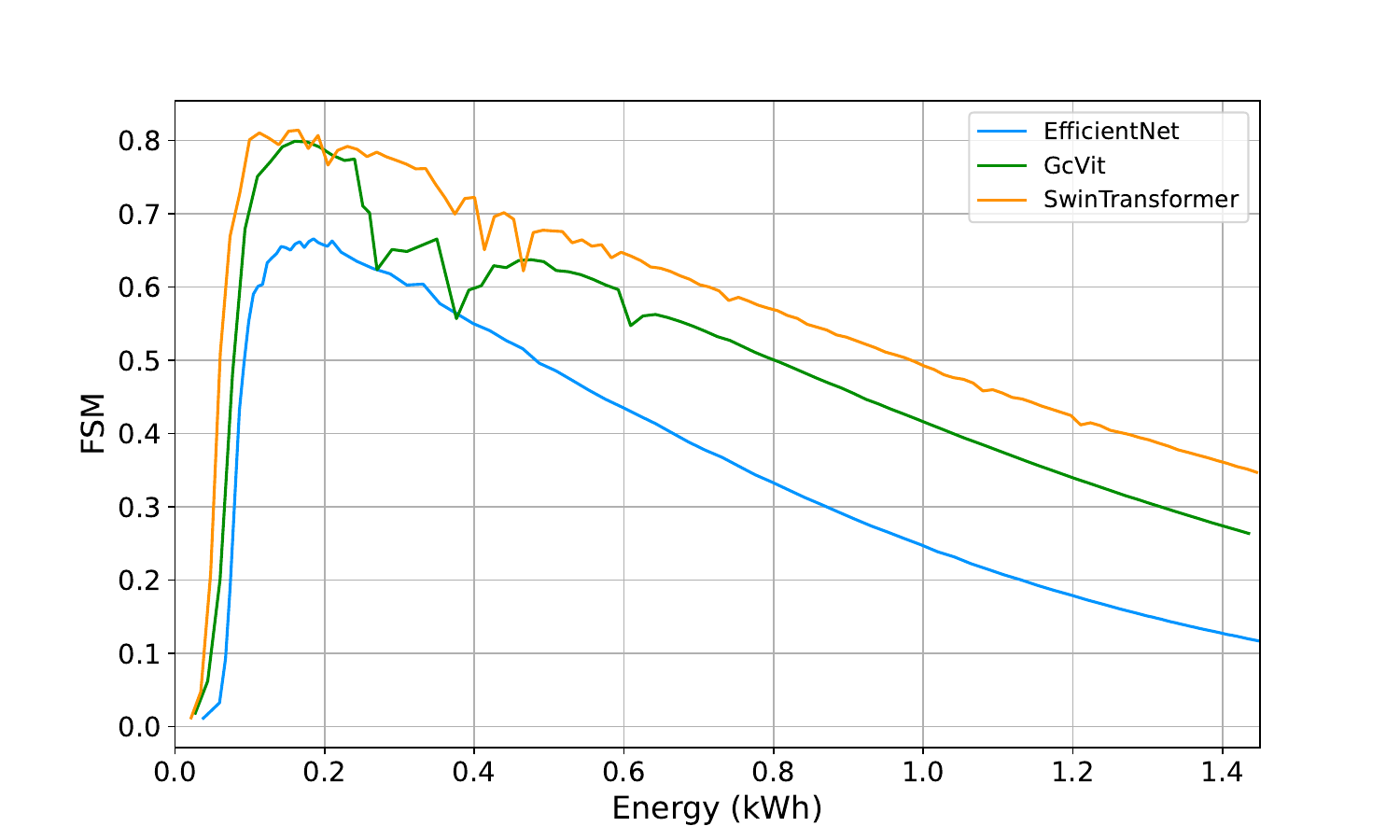}
    \caption{FMS vs Energy on ImageNet for EfficientNet, SwinTransformer, and GcViT.}
    \label{fig:FMS_energy}
    \vspace{-10pt}
\end{figure}
\begin{figure}[t]
    \centering
    \includegraphics[width=0.48\columnwidth]{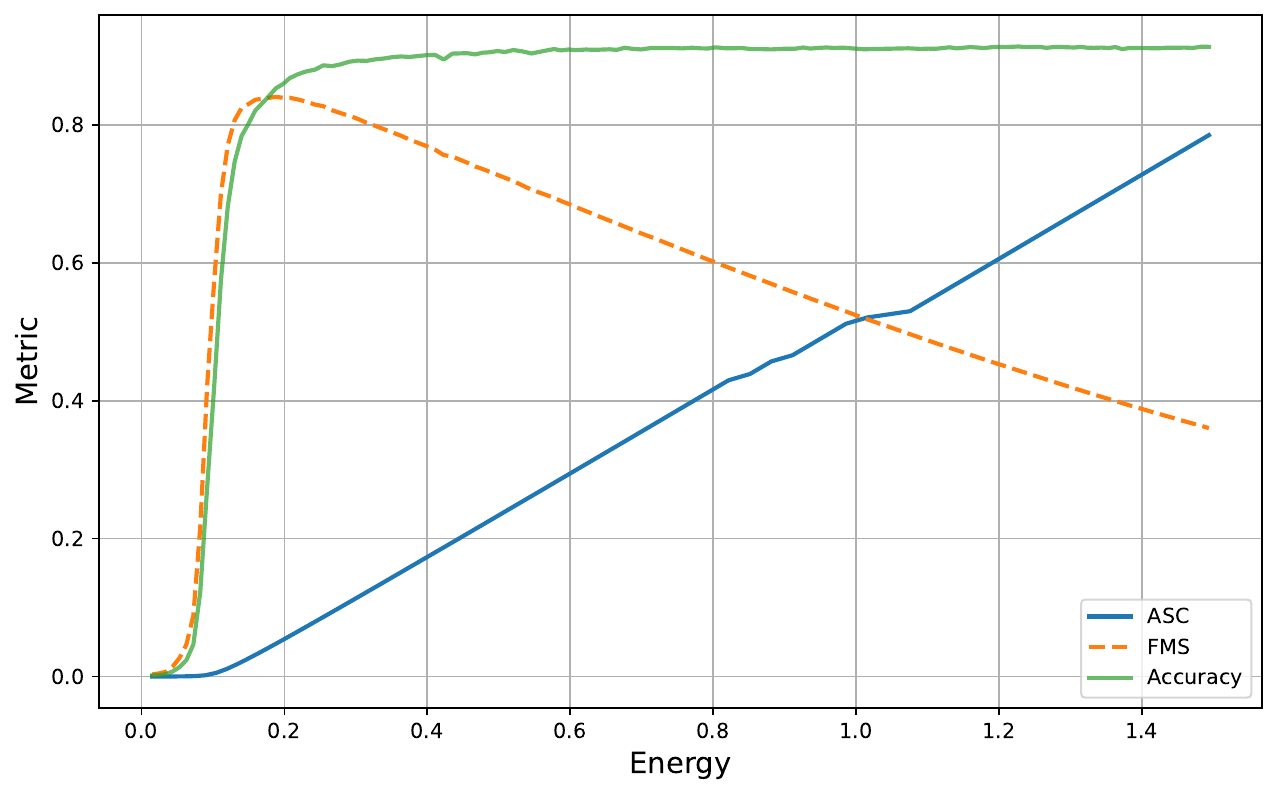}
    \caption{Comparison of accuracy, FMS, and ASC with power values on the GcViT\_tiny dataset}
    \vspace{-13pt}
    \label{fig:comparison_metric}
\end{figure}

\noindent \textbf{Results Analysis}  From Fig. \ref{fig:accuracy_energy} and Fig. \ref{fig:FMS_energy}, we can see that the sustainability metrics of EfficientNet and GcViT are penalized to different degrees due to their respective disadvantages: EfficientNet has low accuracy, and GcViT has high energy consumption. Here, SwinTransformer achieves the highest FMS and ASC on the same dataset by effectively balancing energy consumption and accuracy. At the same time, since the data scale of CIFAR is smaller than that of ImageNet, the FMS and ASC values of each model on ImageNet are smaller than those of CIFAR. These results show the rationality of our proposed criteria, which is consistent with our general understanding of the model. Correspondingly, the previous work SAM  value has a negative value in EfficientNet and is significantly different from the values on the other two models. If we pay attention to the accuracy of EfficientNet, we can find that its accuracy is much lower than that of the other two models. Therefore, the significant difference between the SAM values is not reasonable when considering both accuracy and energy consumption comprehensively. Fig. \ref{fig:comparison_metric} shows that when the power value increases, FMS and accuracy change in opposite directions, while ASC and accuracy change in the same direction, indicating that ASC gives greater weight to accuracy compared to FMS.

\begin{figure}[!htp]
    \centering
    \vspace{-10pt}
    \includegraphics[width=0.57\columnwidth]{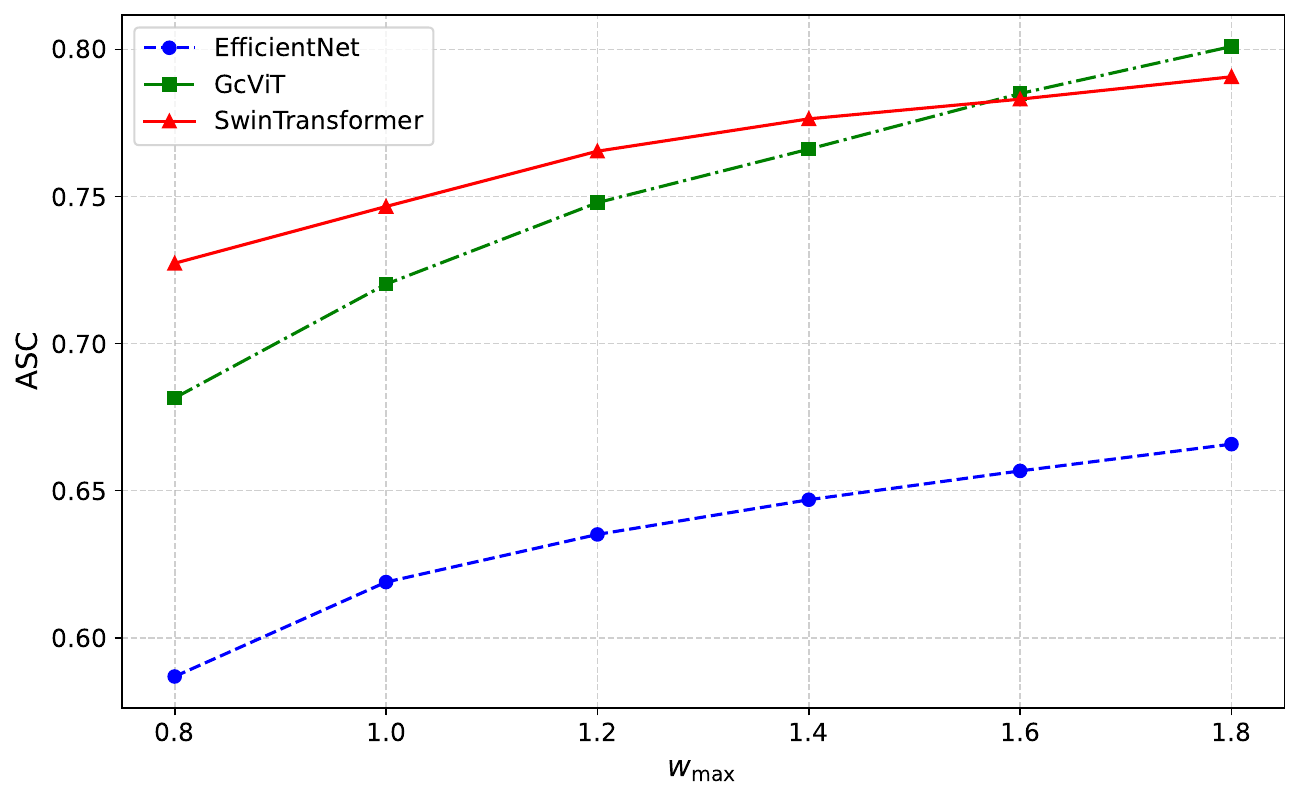}
    \vspace{-10pt}
    \caption{Comparison of ASC values of three models GcViT, SwinTransformer, and EfficientNet under different cutoff power $w_{\max}$}
    \vspace{-8pt}
    \label{fig:diffenergy}
\end{figure}

\noindent \textbf{Ablation Study on Parameter $w_{\max}$ of ASC} From the calculation of ASC, it can be seen that the calculation of ASC value depends on the cutoff power value $w_{\max}$ when the algorithms are compared. Fig. \ref{fig:diffenergy} shows that when $w_{\max}$ increases, the ASC values of the three algorithms increase slowly. The accuracy of the algorithm will gradually increase with the increase of power consumption, so the weighted average of the accuracy (ASC value) will also increase.

It is worth noting that the gradient of the curve measures the marginal gain in performance as power values increase. When we use a larger cutoff power value $w_{\max}$, the GcViT algorithm has a higher marginal gain in performance than the other two algorithms. To some extent, GcViT is less prone to overfitting; that is, as we increase the power value, we can continue to improve the algorithm's performance.

\begin{table}[ht]
    \centering
    \scriptsize 
    \renewcommand{\arraystretch}{1.2} 
    \vspace{-10pt}
    \caption{Comparison of different parameter levels of GcViT on the ImageNet dataset.}
    \label{tab:gcvit_comparison}
    \begin{tabular}{lcccccccc}
        \toprule
        \textbf{Model} & \textbf{Params (M)} & \textbf{TE (kWh)} & \textbf{Acc (\%)} & \textbf{Score} & \textbf{SI} & \textbf{SAM} & \textbf{FMS (\%)} & \textbf{ASC (\%)} \\
        \midrule
        GcViT\_tiny  & \textbf{28.22}  & \textbf{0.46}  & 89.03  & \textbf{1.94}   & \textbf{1.39}  & -8.20  & \textbf{78.23}  & \textbf{83.93} \\
        GcViT\_base  & 90.32  & 1.43  & \textbf{90.40}  & 0.63  & 0.80  & \textbf{19.25}  & 40.07  & 77.02  \\
        \bottomrule
    \end{tabular}
    \vspace{-10pt}
\end{table}

\noindent \textbf{Energy-Performance Trade-off in GcViT} In the ImageNet dataset, we conducted experiments on GcViT\_tiny and GcViT\_base (see Tab.~\ref{tab:gcvit_comparison}). The results indicate that GcViT\_base achieves a slightly higher test accuracy (90.40\%) compared to GcViT\_tiny (89.03\%). However, its training energy consumption (1.4348 kWh) is significantly higher than that of GcViT\_tiny (0.4561 kWh), demonstrating that larger models lead to higher computational costs.  

Moreover, GcViT\_tiny significantly outperforms GcViT\_base in FSM (78.23\% vs. 40.07\%) and ASC (83.93\% vs. 77.02\%), indicating higher feature utilization and computational efficiency, achieving better performance under the same computational resources. Meanwhile, the SAM metric assigns a much higher value to GcViT\_base due to its increased energy consumption, whereas GcViT\_tiny, despite its significantly lower energy consumption, achieves comparable accuracy.  This highlights the limitations of SAM in effectively balancing energy efficiency and performance, whereas FSM and ASC provide more insightful guidance for model evaluation and improvement. By leveraging FSM and ASC as evaluation criteria, model compression and neural architecture search (NAS) can be effectively guided to select more efficient models in resource-constrained environments.

\subsection{Image Segmentation Task}

\noindent \textbf{Experimental Settings} We use our proposed criteria FMS and ASC to evaluate the sustainability of semantic segmentation tasks on the classic Cityscapes dataset \cite{cordts2016cityscapes}, including Deeplabv3\cite{deeplabv3}, MobileNet\cite{mobilenet}, PSANet\cite{zhao2018psanet}, and APCNet\cite{apc}. The specific parameters and results of the models are detailed in Tab.~\ref{tab:model_comparison2}. We set the number of iterations to 80,000 and the batch size to 8. The Cityscapes dataset used for the experiments includes images from the \texttt{leftImg8bit} subset, with a resolution of 2048$\times$1024 pixels for both training and evaluation.

\noindent \textbf{Results Analysis}  It can be seen that MobileNet stands out for its lightweight characteristics, while APCNet exhibits higher parameter complexity and consumes more energy. In addition, we also compare the sustainability values of three different versions of the $setr\_vit\_model$, including naive, mla, and pup versions. The \tf{naive} version adopts a lightweight training strategy based on ViT-Large for fast benchmarking, while \tf{mla} enhances feature representation through a multi-level attention (MLA) strategy to improve segmentation accuracy. In contrast, the \tf{pup} adopts a more complex progressive upsampling (PUP) strategy to restore image details, making it suitable for small object segmentation, but significantly increasing energy consumption. It can be seen that the three evaluation criteria achieve consistent results on the segmentation task, but due to the smaller differences in energy consumption and performance between the models, our criterion function gives a more reasonable degree of discrimination.

\begin{table*}[t]
\centering
\scriptsize 
\renewcommand{\arraystretch}{1.2} 
\caption{Comparison of sustainability criteria of different models on segmentation tasks, where the performance metric is the average of the accuracies of multiple categories (aAcc), and TE (kWh) means train energy consumption.}
\label{tab:model_comparison2}
\vspace{10pt}
\begin{tabular}{lcccccccc}
\toprule
\textbf{Model} & \textbf{Params (M)} & \textbf{aAcc (\%)} & \textbf{TE (kWh)} & \textbf{Score} & \textbf{SI} & \textbf{SAM} & \textbf{FMS (\%)} & \textbf{ASC (\%)} \\
\midrule
MobileNet & \textbf{18.38}  & \textbf{94.57}  & \textbf{0.62}  & \textbf{1.52}  & \textbf{1.24 } & -18.09  & \textbf{92.83}  & \textbf{90.29}  \\
PSANet    & 56.77  & 94.45  & 1.16  & 0.82  & 0.90  & \textbf{59.52}  & 78.71  & 89.66  \\
DeepLabV3 & 65.75  & 93.86  & 1.55  & 0.61  & 0.78  & 19.24  & 56.25  & 87.14  \\
APCNet    & 72.99  & 91.49  & 1.73  & 0.53  & 0.73  & 13.43  & 30.79  & 84.11  \\
\midrule
mla       & 310.68  & \textbf{84.76}  & \textbf{2.52}  & \textbf{0.34}  & \textbf{0.58}  & \textbf{5.46}  & \textbf{11.96}  & \textbf{79.81}  \\
naive     & \textbf{306.54}  & 79.52  & 2.82  & 0.28  & 0.53  & 3.53  & 7.88  & 76.48  \\
pup       & 318.47  & 80.58  & 3.23  & 0.25  & 0.50  & 3.33  & 0.43  & 71.13  \\
\bottomrule
\end{tabular}
\end{table*}

\subsection{Pose Estimation Task}
\noindent \textbf{Experimental Settings} Pose estimation is of great significance for understanding the behavior of organisms, including humans. In our work, we perform a sustainable evaluation on the pose estimation task of wild animals based on the AP-10K dataset. AP-10K\cite{ap-10k} is a large-scale pose estimation benchmark dataset focused on mammals, containing 10,015 images, covering 23 animal families and 54 species. We conduct a comprehensive evaluation of animal pose estimation models based on the task-driven MMPose~\cite{mmpose2020} with two different backbone networks, ResNet50 and ResNet101.

\noindent \textbf{Results Analysis} We iterated the model 30,000 times, and the specific parameters of the model are detailed in Tab.~\ref{tab:model_comparison}. The model based on ResNet-50 has lower computational complexity and higher energy efficiency, making it very suitable for energy-constrained scenarios. In contrast, the model based on the deeper ResNet-101 architecture has significantly higher energy consumption due to the increase in the number of parameters and higher floating-point operations (FLOPs). This comparative experiment also illustrates how we should choose when using different backbone networks, considering the cost of performance improvement.

\begin{figure}[!htp]
    \centering
    \vspace{-25pt}
    \includegraphics[width=1.0\textwidth]{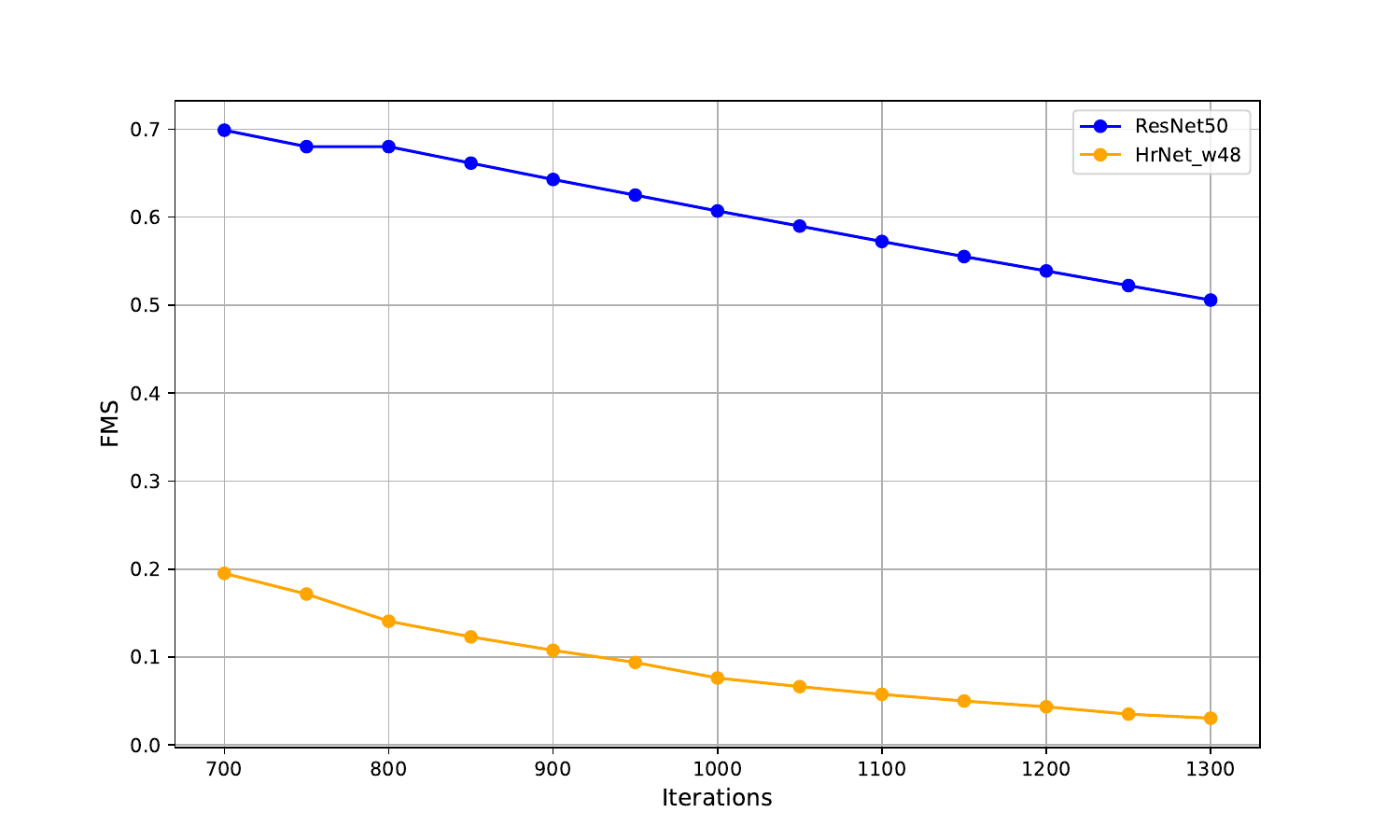}
    \caption{The curve of FMS value changing with different $\alpha$ settings: $\alpha$ is set to 100 times the energy consumption after the $k$-th iteration}
    \label{fig:resnetalpha}
    \vspace{-20pt}
\end{figure}

\noindent \textbf{Ablation Study on Parameter $\alpha$ of FMS} We study the effect of the parameter $\alpha$ on the FMS measure in the pose estimation problem, as shown in Fig. \ref{fig:resnetalpha}. Note that the horizontal axis represents the number of algorithm iterations. For each iteration, the $\alpha$ value is set to 100 times the energy consumption value for that iteration. As the alpha value increases, the FMS on the two models ResNet50 and HrNet\_w48 \cite{HrNet} gradually decreases. It can be seen that the FMS sustainability metric of ResNet50 is always better than that of HrNet\_w48, so different values $\alpha$ do not change the relative order of the comparison.

\begin{table*}[t]
    \centering
    \scriptsize 
    \renewcommand{\arraystretch}{1.2} 
    \caption{Comparison of sustainability criteria of two backbones on pose recognition tasks, where the performance metric is the aAcc, and TE (kWh) means train energy consumption.}
    \vspace{10pt}
    \label{tab:model_comparison}
    \begin{tabular}{lccccccccc}
        \toprule
        \textbf{Model} & \textbf{FLOPs(G)} & \textbf{Params(M)} & \textbf{aAcc(\%)} & \textbf{TE(kWh)} & \textbf{Score} & \textbf{SI} & \textbf{SAM} & \textbf{FMS (\%)} & \textbf{ASC (\%)} \\
        \midrule
        Res50  & \textbf{5.45}  & \textbf{34.00}  & 93.60  & \textbf{0.49}  & \textbf{1.91}  & \textbf{1.38}  & \textbf{-11.58}  & \textbf{61.40}  & \textbf{91.83} \\
        Res101 & 9.10  & 52.99  & \textbf{94.36}  & 0.73  & 1.29  & 1.13  & -27.68  & 31.00  & 91.23  \\
        \bottomrule
    \end{tabular}
\end{table*}

\noindent \textbf{Ablation Study on Parameter $\beta$ of FSM} Below, we discuss the effect of parameter $\beta$ on the sustainability metric FMS, where FMS is defined as follows:
\begin{eqnarray}\label{eqn:fms-d}
\textrm{FMS} & = & (1 + \beta^2) \cdot \frac{P \cdot E}{\beta^2 P + E} \nonumber \\
& = & (1 + \beta^2) \cdot \frac{1}{\beta^2 E^{-1} + P^{-1}} \nonumber \\
& = & \frac{1}{\frac{\beta^2}{1 + \beta^2} E^{-1} + \frac{1}{1 + \beta^2}P^{-1}}.
\end{eqnarray}

We know that the weight $\beta^2/(1 + \beta^2)$ of the energy consumption metric $E$ is an increasing function of the parameter $\beta$. Therefore, when $\beta > 1$, FMS gives a higher weight to the energy consumption metric $E$ than to the performance $P$, that is, a greater penalty is given to high-energy consumption algorithms; when $\beta < 1$, FMS gives a higher weight to the performance $P$ than to the energy consumption metric $E$, that is, a higher reward is given to high-performance algorithms.

\begin{table*}[!htp]
    \small
    \centering
    \caption{Comparison of FMS metric of two algorithms ResNet50 and ResNet101 on pose estimation tasks under different $\beta$ values}
    \label{tab:model_comparison_fms}
    \vspace{5pt}
    \begin{tabular}{lcc}
        \toprule
        \textbf{Metric} & \textbf{ResNet50} & \textbf{ResNet101} \\
        \midrule
        FLOPs (G) & 5.45 & 9.10 \\
        Params (M) & 34.00 & 52.99 \\
        aAcc (\%) & 93.60 & 94.36 \\
        Train Energy (kWh) & 0.49 & 0.73 \\
        E(w)(\%) & 45.68 & 18.54 \\
        FMS ($\beta=0.5$)(\%) & 76.76 & 51.31 \\
        FMS ($\beta=1$)(\%) & 61.16 & 30.86 \\
        FMS ($\beta=2$)(\%) & 50.82 & 22.07 \\
        \bottomrule
    \end{tabular}
\end{table*}

As shown in Tab. \ref{tab:model_comparison_fms}, when $\beta = 0.5$, FMS is closer to the aAcc performance value (more weight to performance) compared to E(w); and when $\beta = 2$, FMS is closer to $E(w)$. ResNet101 consumes more energy, but the performance improvement is limited, resulting in a lower FMS value. Under the three values of $\beta$, the sustainability indicators of ResNet50 are better than those of ResNet101.

\subsection{Batch \& Online Learning}

\noindent \textbf{Experimental Settings} Online learning and batch learning\cite{shalev2011online} are two crucial machine learning methods in machine learning: online models update the model immediately after each sample arrives. In contrast, batch learning updates the model after a batch of samples arrives. Below, we discuss how our sustainable metrics can reasonably evaluate two tasks: two-class and multi-class classification on the CreditCard dataset \cite{dal2015credit} and the ImageSegments dataset \cite{ImageSegments}. For the two-class credit card risk detection application, we utilize the MLP model to evaluate the sustainability of two different implementations. For the multi-class image segmentation application, we compare the sustainability metrics of batch and online decision tree classifiers.
\\
\noindent \textbf{Results Analysis} Generally, online learning consumes more energy because it updates parameters more frequently than batch learning. However, we can observe from Table \ref{tab:comparison7} that the energy consumption of online learning algorithms is significantly higher than that of batch algorithms, but their performance is much higher than that of batch algorithms. In this case, our sustainability metric tends to give online learning a higher score. In general, the sustainability metric will comprehensively consider the differences between the two and give a reasonable score.

\begin{table}[t]
    \centering
    \scriptsize
    \renewcommand{\arraystretch}{1.2} 
    \vspace{-10pt}
    \caption{Comparison of different methods for Binary and Multi-Class tasks.}
    \label{tab:comparison7}
    \resizebox{\textwidth}{!}{
    \begin{tabular}{llccccccc}
        \toprule
        \textbf{Task} & \textbf{Method} & \textbf{E (kWh)} & \textbf{AUC/Acc (\%)} & \textbf{Score} & \textbf{SI} & \textbf{SAM} & \textbf{FMS (\%)} & \textbf{ASC (\%)} \\
        \midrule
        \multirow{2}{*}{Binary} 
            & Batch  & \textbf{0.0086}  & 51.70 (AUC)  & \textbf{60.12}  & \textbf{7.75}  & \textbf{-0.09}  & 68.16  & 49.25  \\
            & Online & 0.5624  & \textbf{69.82 (AUC)}  & 1.24  & 1.11  & -3.32  & \textbf{82.00}  & \textbf{59.98}  \\
        \midrule
        \multirow{2}{*}{Multi-Class} 
            & Batch  & \textbf{0.0009}  & 68.56 (Acc)  & \textbf{761.78}  & \textbf{27.60}  & \textbf{-0.25}  & 80.24  & 66.80  \\
            & Online & 0.2416  & \textbf{79.27 (Acc)}  & 3.28  & 1.81  & -2.54  & \textbf{88.26}  & \textbf{74.05}  \\
        \bottomrule
    \end{tabular}}
    \vspace{-10pt}
\end{table}

\section{Conclusion}

We proposed two novel metrics, FMS (F-Measure on Sustainability) and ASC (Area under Sustainability Curve), to assess the balance between performance and energy efficiency in deep learning models.  FMS combines energy consumption and algorithm performance into a single harmonic mean, while ASC captures the relationship between energy usage and performance by calculating the area under the sustainability curve.  Both metrics are scale-invariant with respect to energy consumption, ensuring their adaptability across diverse scenarios.  Extensive experiments on tasks such as image classification, segmentation, and pose estimation demonstrate their effectiveness, providing a robust and practical framework for comparing algorithms and developing energy-efficient, high-performance models.  Future directions include extending these metrics to broader domains, incorporating hardware-aware optimizations, addressing power distribution and instantaneous demands, and integrating carbon footprint estimation to enhance sustainability considerations.

\bibliographystyle{splncs04}
\bibliography{prcv}

\newpage

\section*{Appendix}
\appendix
\section{Details on  Tasks}
\subsection{Image Classification}

The transform pipeline standardizes the raw images into a format suitable for deep learning models.  Specifically, it resizes each image to 224×224 pixels, which is the standard input size for many popular image classification models, such as ResNet and ViT.  After resizing, the image is converted into a tensor format, followed by channel-wise normalization using the mean and standard deviation derived from the ImageNet dataset (mean=[0.485, 0.456, 0.406], std=[0.229, 0.224, 0.225]).  This preprocessing not only ensures consistent image dimensions and pixel distributions but also helps align the input distribution with that of pretrained models, thereby improving performance during transfer learning.

\subsection{Image Segmentation}

This task employs various data preprocessing strategies in the image segmentation task to enhance the model's generalization ability and computational stability. During the training phase, images from the Cityscapes dataset are first loaded into the model and undergo random scaling (scale range 0.5–2.0, base resolution 2048×1024) to improve adaptability to different object sizes. Subsequently, random cropping is applied (fixed crop size 512×1024) with a category maximum ratio of 0.75 to prevent a single class from dominating the cropped region. Additionally, images are horizontally flipped with a 50\% probability, and photometric distortions (including brightness, contrast, and hue variations) are introduced to enhance data diversity. In the testing phase, a fixed scaling strategy is used to resize images to 2048×1024, maintaining the aspect ratio to ensure the comparability of evaluation results. Furthermore, test-time augmentation (TTA) is implemented to improve model robustness, specifically by applying six scales (0.5, 0.75, 1.0, 1.25, 1.5, 1.75) for multi-scale inference, along with horizontal flipping, to optimize model performance under varying conditions.

\subsection{Pose Estimation}

This study employs a series of data preprocessing and parameter optimization strategies in the pose estimation task to ensure model training stability and inference efficiency. In terms of data preprocessing, PoseDataPreprocessor is used for normalization, specifically applying a mean of [123.675, 116.28, 103.53] and a standard deviation of [58.395, 57.12, 57.375] to reduce the impact of lighting and color variations on the model. Additionally, Batch Normalization is adopted during training to accelerate convergence and ensure consistency across different data batches. For data loading, the batch size is set to 64, combined with 4 data loading workers (num\_workers=4), leveraging multi-threaded parallel processing to improve data loading efficiency and reduce I/O bottlenecks during training. Furthermore, to enhance training stability and reproducibility, a series of Hook mechanisms is configured for training monitoring and anomaly detection, including log recording (every 50 iterations), energy monitoring (Energy Hook, computed every 50 iterations), and anomaly sample detection (Bad Case Analysis, filtering samples with a loss greater than 5).

\section{Ablation Study on Parameter $N$ in ASC}

In the Image Classification Task, we investigate the effect of different values of N on the ASC score. Experimental results show that when N is small, the ASC score tends to favor models with higher accuracy. As N increases, the ASC score shifts its preference toward lightweight models with lower energy consumption. This is because lightweight models often require more training iterations to reach high accuracy, whereas larger models achieve high accuracy early in training but show diminishing improvement over time. Therefore, with a longer evaluation window, the efficiency advantage of lightweight models becomes more apparent, ultimately leading to higher ASC scores. This further demonstrates that ASC can dynamically balance the trade-off between performance and energy efficiency under different evaluation granularities.

\begin{table}[ht]
\centering
\caption{Comparison of ASC values under different $N$ settings on ImageNet and CIFAR datasets.}
\label{tab:asc_comparison_swapped}
\begin{tabular}{lcccccc}
\toprule
\multicolumn{4}{c}{\textbf{ImageNet}} & \multicolumn{3}{c}{\textbf{CIFAR}} \\
\midrule
\textbf{N} & \textbf{EfficientNet} & \textbf{SwinTransformer} & \textbf{GcViT} 
& \textbf{EfficientNet} & \textbf{SwinTransformer} & \textbf{GcViT} \\
\midrule
10  & 0.6658 & 0.7907 & \textbf{0.8010} 
    & 0.6011 & \textbf{0.8124} & 0.6703 \\
50  & 0.6537 & 0.7851 & \textbf{0.7934} 
    & 0.6008 & \textbf{0.8115} & 0.6699 \\
100 & 0.5783 & \textbf{0.7862} & 0.7858 
    & 0.6008 & \textbf{0.8113} & 0.6699 \\
500 & 0.2082 & \textbf{0.2171} & 0.1724 
    & \textbf{0.6006} & 0.4215 & 0.3490 \\
\bottomrule
\end{tabular}
\end{table}

\section{Ablation Study of Performance Metric in Sustainability Metric}
In the sustainability metric, performance can be any performance measure value for the task on the test set. Tab.~\ref{tab:model_comparison_swapped} shows the results of using two different performance measures aAcc and mIoU on the segmentation task. In essence, these different performance measures represent different goals of the algorithm from different perspectives. Therefore, for the sustainability of the algorithm, we will get different conclusions based on different goals. For example, according to the ASC measure, we use aAcc and mIoU performance measures to deduce that MobileNet and PSANet are the best among the four algorithms, respectively.

In general, the ranking results of algorithm comparison depend on the goals we pursue.

\begin{table*}[ht]
\centering
\caption{Comparison of sustainability criteria of two backbones on pose recognition tasks, where the performance metric is the aAcc, and TE(kwh) means train energy consumption.}
\label{tab:model_comparison_swapped}

\begin{tabular}{lcccccc}
\toprule
\textbf{Metric} & \textbf{Model} & \textbf{Performance(\%)} & \textbf{Score} & \textbf{SAM} & \textbf{FMS(\%)} & \textbf{ASC(\%)} \\
\midrule
\multirow{4}{*}{aAcc} 
    & MobileNet & \textbf{94.57} & \textbf{1.52} & -18.09 & \textbf{92.83} & \textbf{90.29} \\
    & PSANet    & 94.45 & 0.82 & \textbf{59.52} & 78.71 & 89.66 \\
    & DeepLabV3 & 93.86 & 0.61 & 19.24 & 56.25 & 87.14 \\
    & APCNet    & 91.49 & 0.53 & 13.43 & 30.79 & 84.11 \\
\midrule
\multirow{4}{*}{mIoU} 
    & MobileNet & 67.02 & \textbf{1.08} & -3.23 & \textbf{77.24} & 52.59 \\
    & PSANet    & \textbf{69.02} & 0.60 & \textbf{12.40} & 63.42 & \textbf{57.74} \\
    & DeepLabV3 & 65.13 & 0.42 & 3.10 & 49.68 & 45.31 \\
    & APCNet    & 58.62 & 0.34 & 1.45 & 29.05 & 42.22 \\
\bottomrule
\end{tabular}
\end{table*}
\end{document}